# Using Natural Language Processing to Screen Patients with Active Heart Failure: An Exploration for Hospital-wide Surveillance.

**Shu Dong, MS[1], R Kannan Mutharasan, MD[2], Siddhartha Jonnalagadda, PhD[1]**

[1]Division of Health and Biomedical Informatics, Department of Preventive Medicine, Northwestern University Feinberg School of Medicine, Chicago, IL
[2]Division of Cardiology, Department of Medicine, Northwestern University Feinberg School of Medicine, Chicago, IL

**Abstract**

*In this paper, we proposed two different approaches, a rule-based approach and a machine-learning based approach, to identify active heart failure cases automatically by analyzing electronic health records (EHR). For the rule-based approach, we extracted cardiovascular data elements from clinical notes and matched patients to different colors according their heart failure condition by using rules provided by experts in heart failure. It achieved 69.4% accuracy and 0.729 F1-Score. For the machine learning approach, with bigram of clinical notes as features, we tried four different models while SVM with linear kernel achieved the best performance with 87.5% accuracy and 0.86 F1-Score. Also, from the classification comparison between the four different models, we believe that linear models fit better for this problem. Once we combine the machine-learning and rule-based algorithms, we will enable hospital-wide surveillance of active heart failure through increased accuracy and interpretability of the outputs.*

**Introduction**

Reducing heart failure readmissions remains a major target for improving quality of care and reducing healthcare costs.[1] A major component of this is identification of heart failure cases upon hospitalization. Early identification of patients with heart failure allows for early deployment of clinical resources to improve quality of care, continuity of care, and potentially reduce readmissions. The traditional model of accessing specialist care in hospitals relies upon placing a consultation request. Drawing from the operations management literature, this can be termed a "push" model of care delivery.[2] A hallmark of push operations is operational delay, and potential inaccuracies in identifying patients who could benefit from more advanced resources. These flaws reduce health care quality and patient experience.

In contrast, hospital-wide surveillance by a central heart failure-focused team enables a "pull" model of specialist care and multidisciplinary intervention delivery. In a disease surveillance model, a disease-focused team specifically screens hospital admissions for the presence of that disease state, and calls front-line providers to offer specialist care. A pull model of consultation and provision of multidisciplinary care may reduce time lags and increase accuracy, thereby improving quality of care.

Automating identification of patients with heart failure admissions is possible by querying electronic health records (EHR). The use of discrete data from EHR is well established as a strategy for identification of disease states.[3] The use of only discrete data, however, circumscribes the efficacy of automated detection methods by neglecting the rich clinical information embedded in free text such as clinical notes, radiological reports, and nursing notes. Natural language processing (NLP) is a strategy whereby clinical data can be mined from free text and meaningfully parsed into a format tractable for further processing by computational algorithms that enable machine learning. NLP, or Information extraction (to be precise) is the discovery by computers of new, previously unfound information, by automatically extracting information from different written resources.[4] Earlier attempts were predominantly dictionary- or rule-based systems; however, most modern systems use supervised machine learning where a system is trained to recognize mentions in text based on specific (and numerous) features associated with the mentions that the system learns from annotated corpora. Unstructured information occurs in a wide variety of formats such as "calculated biplane LV ejection fraction is 37%" or "left ventricular function is severely reduced." At a fundamental level, a medical concept such as a disease, treatment, or diagnostic test could be mentioned as a noun phrase – an incomplete sentence (ex: dry mucous membranes and myotic pupils), a complete sentence (ex: The patient had patent carotids bilaterally on her neck MRA), or as a list (ex: Tylenol 650 mg p.o. q. 4-6h p.r.n. headache or pain; acyclovir 400 mg p.o. t.i.d;). Despite active interest from clinicians in its potential,[5] clinical information extraction is an unsolved problem.

Here, we propose a natural language processing-based strategy for automated detection of heart failure cases, with high sensitivity and specificity, sufficient to enable operational use in a clinical setting.

**Methods**

**Problem**

We identified 3867 patients admitted to Northwestern Memorial Hospital between August and September 2015 with a possible diagnosis of heart failure. This initial screen was accomplished with an enterprise data warehouse query to identify patients with possible acute heart failure exacerbation. This screen primarily detects patients on the basis of receiving intravenous diuretic medications (used to remove fluid from patients with congestion), or with blood tests suggestive of heart failure (brain naturetic peptide greater than 100 ng/dl). The results of this EDW query were manually classified by a cardiologist and a nurse clinician, both expert in identification of patients with possible heart failure. Depicted in Table 1 is the classification schema. Although we have five colors here, in our algorithm, we treat Grey, Red and Purple as one class "Other". Our goal here is to classify different patients to these three classes automatically by analyzing the EHR data automatically retrieved from the Northwestern Medicine Enterprise Data Warehouse (NMEDW).

**Table 1: Color to Condition Mapping**

| Color | Description |
|---|---|
| Green | Patient has active heart failure, but no cardiology service is consulted |
| Orange | Patient has active heart failure, however cardiology has been consulted or the patient is on Galter 10 |
| Other(Grey) | Patient does not have heart failure |
| Other(Red) | Patient have a history of heart failure, however that is not the active issue |
| Other(Purple) | Patient has/had a heart transplant/ventricular assist device (VAD) |

**System Overview**

To decouple the system into simpler parts, we split our classification system into four components – ***Profile Builder***, ***Data Element Extractor***, ***Patient Classifier*** and ***Metrics Calculator***. The system architecture is as shown in Figure 1.

We implemented two different classifiers – a rule-based classifier and a machine learning-based classifier. For the two classifiers, we only needed to change the implementation of the second and third components in their pipelines. That is, for the rule-based approach, we extracted data elements first and performed the matching of patients to colors by specific rules. For the machine learning-based approach, we built features for all patients' notes to build a feature matrix and then trained several models against this feature matrix.

The overall workflow works could be described as follows:

1. The ***Profile Builder*** parses the raw data to patient profile, which is an aggregated JSON containing all his/her information. With this profile, we will be able to do analysis against each patient.
2. For the rule-based approach, the ***Data Elements Extractor*** extracts data elements from patient profile and then ***Rule-Based Classifier*** performs the classification according to extracted data elements.
3. For the machine learning based approach, the ***Feature Matrix Builder*** builds the feature matrix from patient profile and then ***Machine Learning-Based Classifier*** is trained to perform the classification.

4. The *Metric Calculator* computes the metrics, including accuracy, recall, precision and F1-score according to classification result.

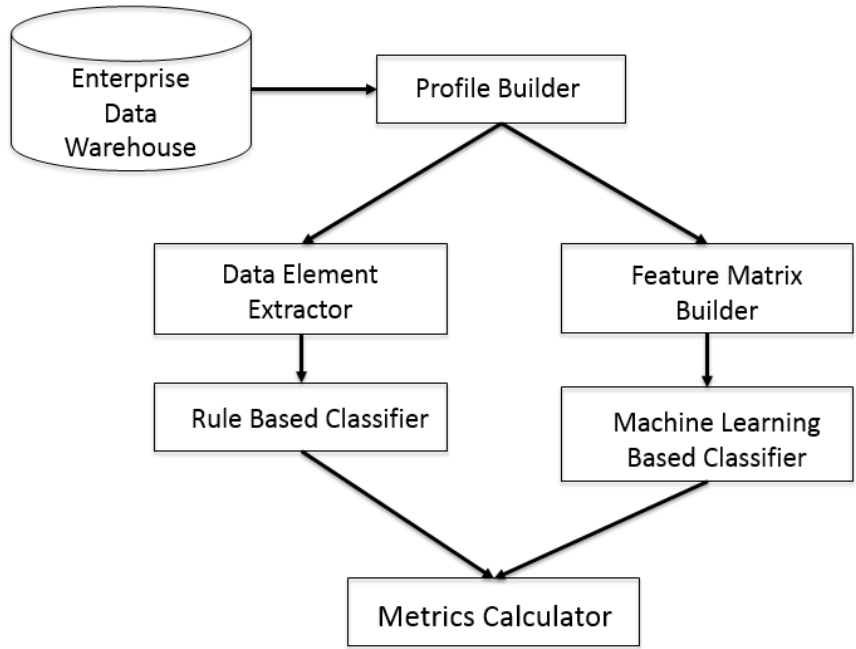

**Figure 1: System Overview**

**Profile Builder**

The raw data retrieved from the NMEDW are clinical notes generated by various doctors, nurses, and allied health professionals caring for a patient during a given hospitalization. Since we are doing the analysis against each patient, we built a profile for each patient by aggregating all their information into a JSON file containing all clinical notes of the patient.

**Rule Based Approach**

The rule-based performs the classification by extracting proper data elements from patients' notes and matching the data elements to given classes according some rules. The intuition of this approach is simulating clinicians to do the choices.

**Table 1: Data Elements**

| Name | Type | Description | Keywords |
|---|---|---|---|
| Cardiology Consulted | Boolean | To indicate if the patient has been cardiology consulted | cardiology, cardiologist |
| Heart Failure | Boolean | To indicate if the patient has Heart Failure or not | HF, SHF, DHF, CHF, heart failure, etc. |
| At Galter 10 | Boolean | To indicate if the patient is at Galter 10 unit | Galter 10 |
| Heart Transplant | Boolean | To indicate if the patient has heart transplant/ventricular assist device | oht, vad, orthotopic heart transplant, heart transplant |
| Non Active Issue | Boolean | To indicate if the patient has non-active Heart Failure | euvolemic |

**Data Elements.** Deciding data elements to be extracted needs a lot of domain knowledge. Table 1 shows the data elements we extracted to feed into classifier, which are five Boolean variables and will be feed into the rule-based classifier. These data elements are all compiled by experts in heart failure. Notice that we first extract keywords listed in the table, and then generate the data elements according to the keywords. For example, if we extracted keywords "cardiology" or "cardiologist", we set the variable "Cardiology Consulted" to be "True", otherwise "False".

For extracting these keywords, we defined regular expressions[6] to match these keywords in the free text clinical notes. However, it is possible that we include some negated keywords during extraction. E.g. "He *denies* any symptomatic precipitants". Simple regex will not work for this case. So, we used NegEX[7] to detect and exclude keywords in such negated statements.

**Classification Rules.** When we have the data elements in structured form, we feed them into the rule-based classifier as shown in Figure 2. The classification rules are also recommended by experts in heart failure. This flowchart aims to classify patients to different groups by imitating the thought process of clinicians.

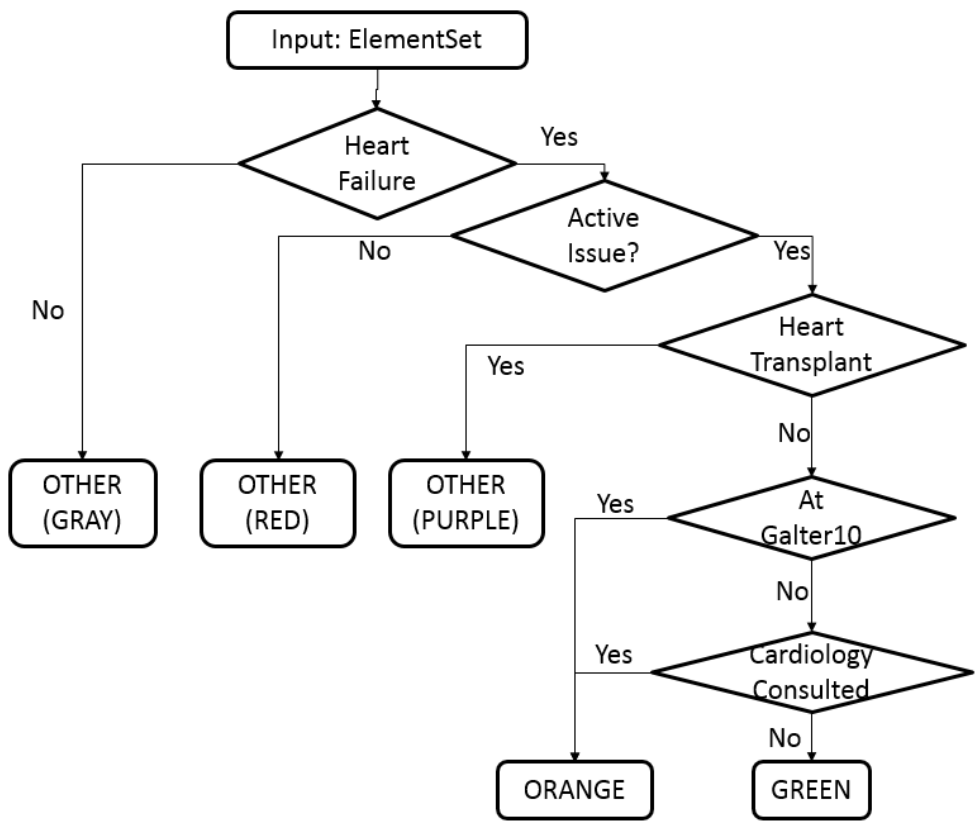

Figure 2: Rule Based Classifier

**Machine Learning-Based Approach**

**Features.** Feature selection[8-10] is very important in machine learning. In this approach, we experimented with several features including n-grams and UMLS concepts derived from cTAKES,[11] etc. for each of the clinical notes in addition to the structured fields for each patient. Before extracting the features, we also performed preprocessing steps such as removing stop words, removing short words with less than four characters and replacing all numbers with a placeholder "NUM". Then, we build feature matrix for each patient against these features.

**Model Training**. We used 10-fold cross validation to train and evaluate the classifiers at a patient-level. In this approach, we picked four machine learning models, including Naïve Bayes,[12] SVM with RBF Kernel, SVM with Linear Kernel,[13] and Logistic Regression,[14] to test their performance and then selected the first one. All the classifiers are implemented by scikit-learn,[15] an open source machine learning library.

There are two problems that need attention in this part. First, the data for each class is not balanced. Figure 3 shows the distribution for each class in our whole data set. From the figure, we can see that 82.3% of the patients belong to "Others" while the "Green" ones only count for 6.1%. To fix this problem, when we were training the model, we set a class weight for each class according to this data distribution. This effectively is equivalent to the oversampling technique[16] used in machine learning for unbalanced training data. Second, the SVM and Logistic Regression models are not capable of classifying multiple classes directly. In this paper, we use the One-vs-Rest strategy,[17] training a single classifier per class, with the samples of that class as positive samples and all other samples as negatives.

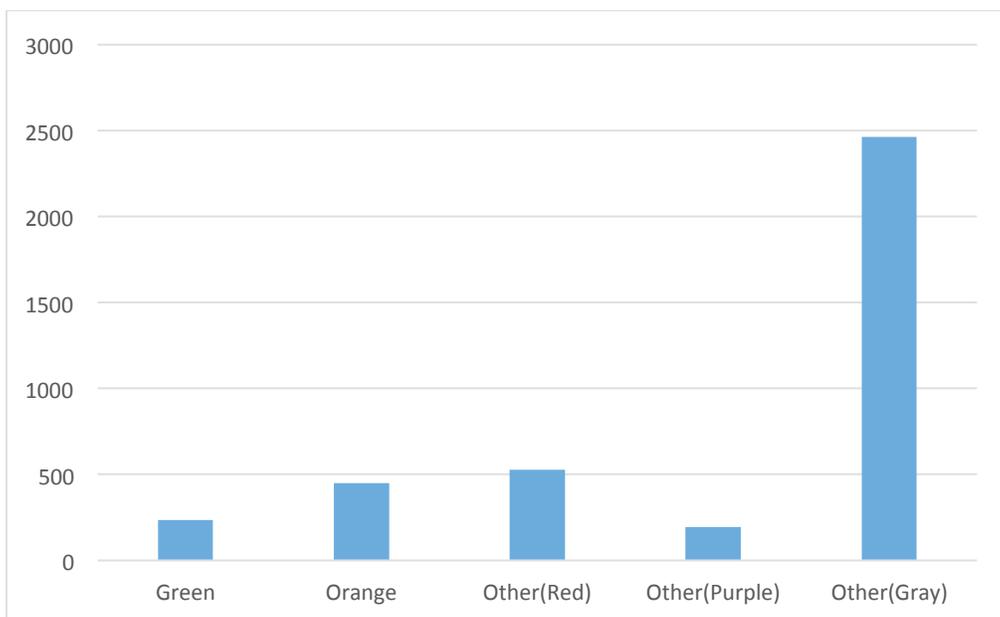

**Figure 3: Input Data Distribution**

**Results**

**Rule-based Classification:** Table 2 shows the classification result for rule-based classifier. We ran our algorithm against all 3834 patients and achieved 69.4% accuracy. However, we believe that with more domain knowledge the rule-based classifier will significantly improve.

**Table 2: Rule Based Classification Result**

| Class | Precision | Recall | F1-score | Accuracy |
|---|---|---|---|---|
| Green | 0.162 | 0.513 | 0.246 | N/A |
| Orange | 0.405 | 0.689 | 0.510 | N/A |
| Other | 0.954 | 0.708 | 0.812 | N/A |
| Avg/Total | 0.830 | 0.684 | 0.729 | 0.694 |

Here[1], for each class, we compute the count of True Positive (TP), True Negative (TN), False Positive (FP) and False Negative (FN) patients for each color first, and then compute the metrics as follows:

$$Precsion = \frac{TP}{TP + FP}$$

$$Recall = \frac{TP}{TP + FN}$$

$$F1 - Socre = 2 \cdot \frac{Precision \bullet Recall}{Precision + Recall}$$

$$Accuracy = \frac{\sum_{color} TP}{\sum_{color} TP + \sum_{color} TN + \sum_{color} FP + \sum_{color} FN}$$

**Machine Learning-based Classification**: Among all text-based features, we found that bigrams of patients' notes with document frequency between 0.2 and 0.8 as data elements produce models with higher accuracy. The bigram with lesser frequency tend to be outliers or spelling mistakes and those with higher frequency tend to be phrases such as "consulting physician" that appear in most notes and not have any predictive power. There are 1816 such bigrams in our feature-set.

Figure 4 shows the comparison of classification result between four models. We compared ***Precision***, ***Recall***, ***F1-Score*** and ***Accuracy***. From the figure, we can see that SVM with Linear Kernel has the best performance in all four parameters, while logistic regression is the second one. We believe that linear models fit better for this problem.

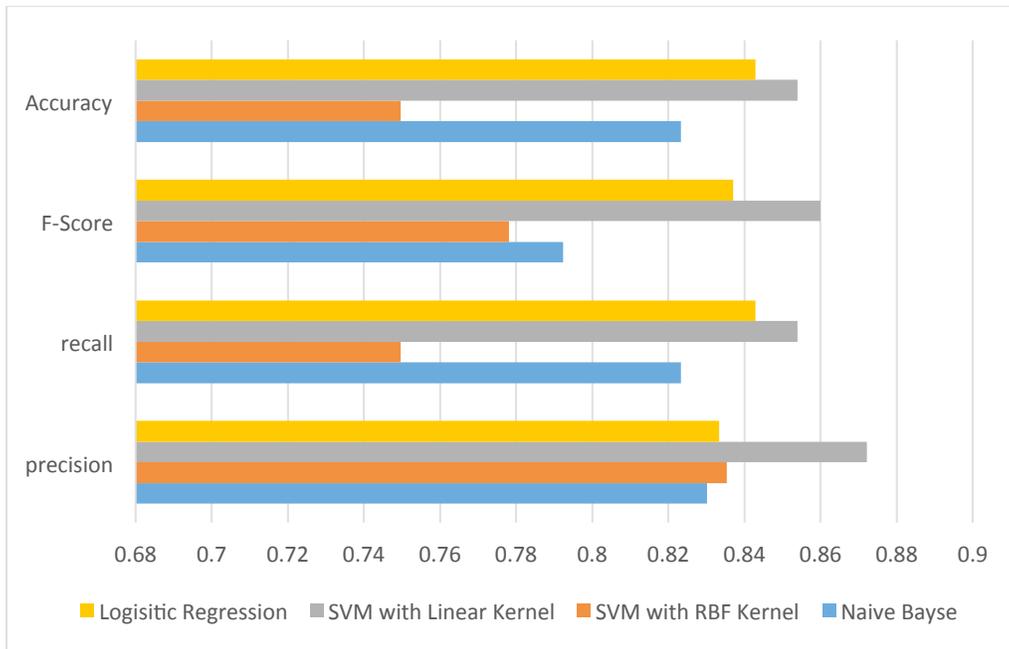

**Figure 4: Model Comparison**

Table 3 shows the detailed classification result using SVM with linear kernel, which outperformed all other models for each of the four metrics.

Table 3: Classification Result of SVM with Linear Kernel

| Class | Precision | Recall | F1-Score | Accuracy |
|---|---|---|---|---|
| Green | 0.54 | 0.49 | 0.51 | N/A |
| Orange | 0.54 | 0.77 | 0.63 | N/A |
| Other | 0.95 | 0.90 | 0.92 | N/A |
| Avg/Total | 0.87 | 0.85 | 0.86 | 0.85 |

**Discussion**

We can make several important observations from the classification results of rule-based approach and machine learning-based approach. For the rule-based approach, precision of green class (Patient has active heart failure, but no cardiology service is consulted) is low since a lot of gray color records (Patient does not have heart failure) are misclassified as green. This is intentional in the design of the algorithm because clinically it is more important to capture all the heart failure cases, even at the expense of misclassifying non heart failure cases as active heart failure. Accuracy can be improved by further work in identifying words in documentation that represent heart failure, as well as extraction of concepts that mark for acuity of the condition. Furthermore, prediction is likely to increase significantly with the improvement and better use of NLP algorithms for context detection in better identifying whether heart failure mentions in text is active, historical or hypothetical.[18] General purpose algorithms such as Context[18] are not accurate enough and need to be adapted for our clinical notes. We will also need to use temporal resolution algorithms[19] to identify the order and timing of events to ascertain whether a heart failure mention is active or not.

From Table 3, we can see that machine-learning based approach achieves a better performance than rules-based approach, achieving 85% accuracy and 0.86 as F1-Score. It is able to identify to the patients that do not need to be screened for active heart failure with a high precision (95%) and high recall (90%). That is, if we use this system across Northwestern Medicine from which this sample is drawn, based on Figure 3 which shows that 82.3% of patients are classified as not heart-failure related, it is reasonable to believe that we will be able to automatically weed out 90% * 82.3% = 74.1% of patients from having to screen for active heart failure. The risk that 5% of the patients screened as not having active heart failure might have active heart failure has operational implications, but we are working with clinicians to render the predictions interpretable and also remove the features that are unrelated and might be leading to overfitting. For example, Table 4 shows the bigrams that the system found to be more predictive. Several of these bigrams such as "health care" are not relevant for the task and will be removed in this process.

Table 4: Most predictive bigrams for SVM with Linear Kernel

| post op | health care | soft non | htn hl | cp sob | replete lytes | function normal |
|---|---|---|---|---|---|---|
| general diet | focal deficits | left atrium | bowel regimen | hypertrophy present | pulm ctab | subcutaneous injection |
| output ml | diastolic function | injection bid | htn hypertension | dysfunction present | extremities warm | application topical |
| daily hs | ct chest | blood loss | aortic root | wheezes crackles | edema neurologic | trace tricuspid |
| po intake | push hours | docusate senna | medications hydromorphone | distended non | high risk | peripheral edema |

**Conclusion**

Based on our result, we conclude that machine learning based approach has a better performance than rule-based approach, with 87.5% accuracy compared to 69.4%. However, our implementation of rule-based approach is not complete. We believe that it could be enhanced by classification rules learnt from machine learning algorithms. When combined, both the algorithms together will not only improve the results to make sure the system will be useful for hospital-wide surveillance of active heart failure, but will also make the classifications interpretable.

**Acknowledgments**

We acknowledge support from the National Library of Medicine (grant R00LM011389 ) and the Bluhm Cardiovascular Institute (BCVI). Several BCVI clinicians and staff funded by them including Kalpana Raja, Corrine Benacka, Robin Fortman, Daniel Navarro, Preeti Kansal, Lynne Goodreau, and Hannah Alphs Jackson have assisted with data preparation and interpretation.